\documentclass[conference]{IEEEtran}
\IEEEoverridecommandlockouts
\usepackage{multirow} 
\usepackage{amsmath}
\usepackage{amssymb}
\usepackage{graphicx}
\usepackage{float}
\usepackage{array}
\usepackage{tikz}
\usepackage{listings}
\usepackage{mathrsfs}
\usepackage{pgfplots}
\usepackage{python}
\usepackage[pdfencoding=auto,psdextra,pagebackref,breaklinks,colorlinks]{hyperref}
\usepackage{graphicx}
\usepackage{subcaption}
\usepackage{cite}
\usepackage{dsfont}
\usepackage{mathtools,etoolbox}
\usepackage{algorithm}
\usepackage{algpseudocode}

\usepackage[none]{hyphenat}
\usepackage[utf8]{inputenc}
\usepackage[english]{babel}

\usepackage{amsthm}
\usepackage{xfrac}
\usepackage{nicefrac}
\usepackage{xcolor}
\usepackage{booktabs}
\usepackage[switch, pagewise]{lineno}
\usepackage{nicefrac}
\usepackage{flushend}
\usepackage{gensymb}
\usepackage{lipsum}
\usepackage{balance}

\def\BibTeX{{\rm B\kern-.05em{\sc i\kern-.025em b}\kern-.08em
    T\kern-.1667em\lower.7ex\hbox{E}\kern-.125emX}}
\begin{document}

\title{RAFT - A Domain Adaptation Framework for RGB \& LiDAR Semantic Segmentation \\
}
\author{
Edward Humes$^{a}$,
Xiaomin Lin$^{a,b}$,
Boxun Hu$^{a}$,
Rithvik Jonna$^{a}$,
and Tinoosh Mohsenin$^{a}$%
\thanks{$^{a}$ Department of Electrical and Computer Engineering, Johns Hopkins University, Baltimore, MD 21218, USA. E-mail: bhu29@jhu.edu}%
\thanks{$^{b}$ Department of Electrical Engineering, University of South Florida, Tampa, FL 33620, USA.}%
}

\makeatletter
\addtocounter{figure}{-2} 
\g@addto@macro\@maketitle{
\begin{figure}[H]
  \setlength{\linewidth}{\textwidth}
  \setlength{\hsize}{\textwidth}
    \centering
    \includegraphics[width=0.95\textwidth]{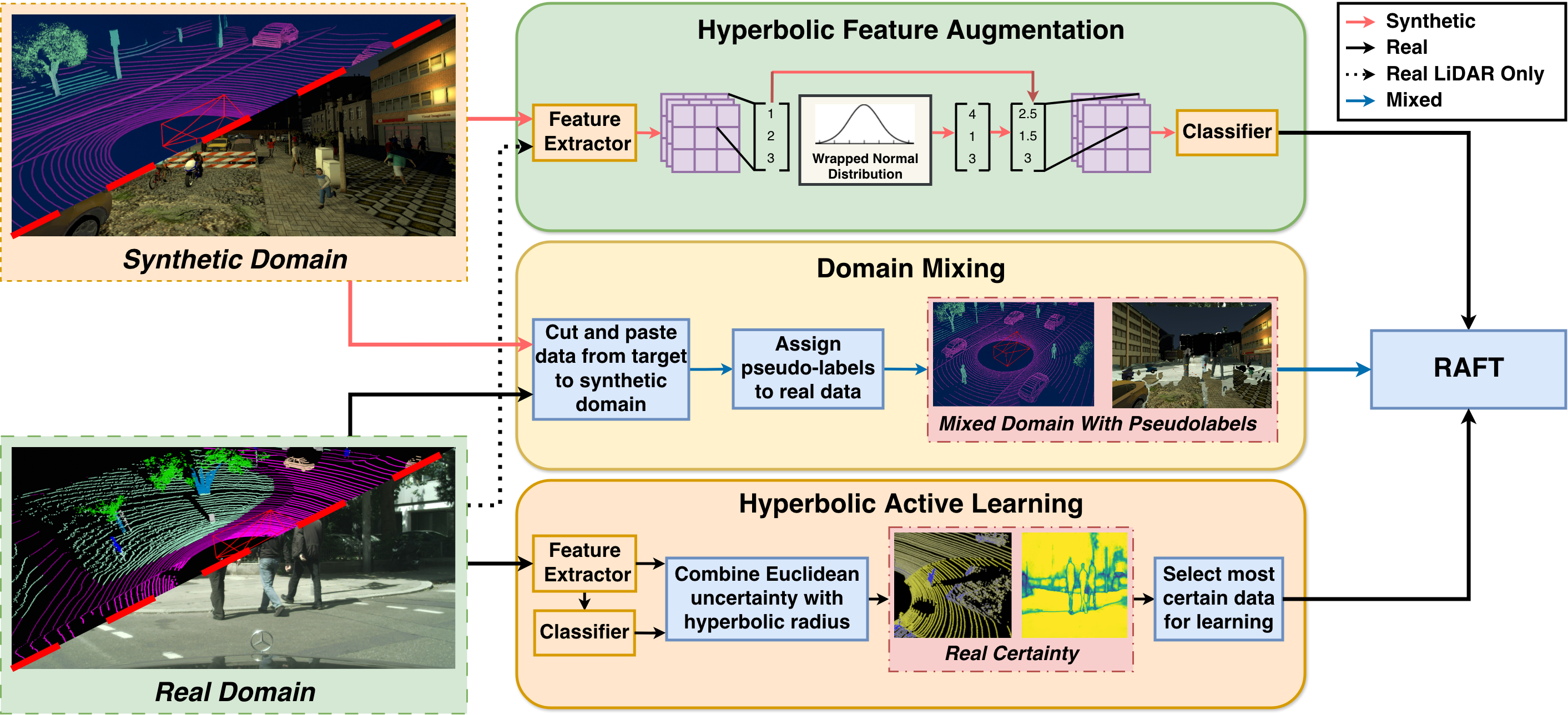}
    \captionsetup{font={footnotesize},labelfont=bf}
    \caption{The proposed architecture of our RAFT framework: it makes use of fully labeled source domain data, and sparsely labeled target domain data acquired via Hyperbolic Active Learning. Hyperbolic Feature Augmentation (HFA) generates novel features to further increase data diversity, and Domain Mixing mixes inputs to create hybrid training examples. The adapted model can then be utilized for downstream perception tasks.}
    \vspace{-5mm}
    \label{fig:proposed_architecture}
    \end{figure}
}

\maketitle
\thispagestyle{empty}
\pagestyle{empty}

\begin{abstract}
Semantic segmentation of RGB images and LiDAR point clouds is central to robotic perception, yet dense real-world labels are costly and models trained on synthetic data face a Syn2Real gap. We introduce \textbf{RAFT} (Robust Augmentation of FeaTures), a unified framework that builds on active domain adaptation and makes the most of a fixed annotation budget by leveraging hyperbolic geometry, extending hyperbolic feature augmentation from image classification to dense prediction, and applying modality-specific domain mixing. Concretely, RAFT uses hyperbolic signals to guide selection and synthesis, generalizes HFA to pixel/voxel embeddings for class-faithful feature-space augmentation, and employs DACS (RGB) or PolarMix (LiDAR) to better align representations. Across SYNTHIA/GTAV $\rightarrow$ Cityscapes and SynLiDAR $\rightarrow$ SemanticKITTI/POSS, RAFT consistently improves over strong ADA baselines while using only a small fraction of target labels - for example, +2.1 mIoU* with SegFormer-B4 on SYNTHIA $\rightarrow$ Cityscapes and +3.3 mIoU with MinkNet on SemanticKITTI. Finally, we validate our adapted models in the real world on a Unitree Go2 equipped with a Livox Mid-360. Furthermore, we release a small, annotated dataset from this sensor to facilitate domain adaptation on this sensor.\url{https://sites.google.com/view/raft-ada-framework/home}.
\end{abstract}



\section{Introduction}



Deep learning–driven semantic segmentation provides scene context~\cite{vinodkumar2023survey, su2021data, guan2022ga} for robotics by classifying every pixel or point and guiding downstream planning. However, the remarkable performance of these models is based on a large amount of meticulously annotated training data~\cite{chen2014big}. Acquiring such datasets for real-world robotics is a major bottleneck~\cite{cordts2016cityscapes, liao2022kitti}, as manual labeling is often expensive, time consuming, and difficult to scale.


Synthetic data offers scale and perfect labels~\cite{lin2022oysternet}, yet models trained on it degrade in the real world due to the Syn2Real domain gap~\cite{hu2023simulation} - differences in noise, lighting, texture, and geometry. For safety-critical applications, the resulting segmentation errors can lead to catastrophic failures.

Active domain adaptation (ADA) aims to bridge this gap by querying a small set of informative target labels. For RGB segmentation: HALO~\cite{franco2023hyperbolic}, and for LiDAR: Annotator~\cite{xie2023annotator}, are state of the art, adapting synthetic pretraining to sparsely labeled target data. While active learning is extremely effective at selecting informative training examples for domain adaptation, it can only do so according to a fixed labeling budget, thus leaving rare classes and edge cases underrepresented. Therefore, we seek to expand the learned distribution as much as possible through input and feature-space augmentations, reserving the limited active learning budget for the truly difficult class instances and rare edge-case scenarios. Concretely, we strengthen the representations generated by the feature extractor, and utilize hyperbolic feature augmentation~\cite{gao2022hyperbolic} to expand class boundaries in feature space so that acquisition steps are spent where labels matter most.

Motivated by this idea, we introduce \textbf{R}obust \textbf{A}ugmentation of \textbf{F}ea\textbf{T}ures (RAFT), a framework that builds on ADA to proactively manufacture hard, label-efficient training signals and close the Syn2Real gap across RGB and LiDAR. As shown in Figure~\ref{fig:proposed_architecture}, RAFT operates entirely in hyperbolic space for both selection and synthesis: hyperbolic radius supplies a principled scarcity/uncertainty signal for active labeling (as in HALO), while the same geometry enables class-faithful synthesis via geodesic interpolation from learned per-class distributions. Coupled with modality-specific domain mixing (DACS for RGB, PolarMix for LiDAR) and hyperbolic feature augmentation (HFA), this dual use expands coverage of long-tailed classes, reduces uncertainty, and respects the data manifold - yielding robust real-world deployment with minimal labeling.

Our key contributions are summarized as follows:
\begin{itemize}
    \item We propose a unified domain adaptation framework for RGB images and LiDAR point clouds that overlays active domain adaptation with targeted input- and feature-space augmentations, and ablate each component.
    \item We extend Hyperbolic Feature Augmentation from classification to pixel and point/voxel-level segmentation, synthesizing rare/ambiguous features that explicitly target edge cases.
    \item We demonstrate state-of-the-art synthetic-to-real performance on standard benchmarks for both modalities while using only a small fraction of real-world annotations.
    \item We validate real-world applicability of the RAFT framework by deploying adapted models on a legged robot, and show improved perception quality in outdoor environments.
    \item We open source a small dataset from the Livox Mid-360 LiDAR(200 training scans, 51 for testing), annotated according to SemanticPOSS' labels and format.
\end{itemize}
\section{Related Works}
Data imbalance in image segmentation arises when dominant regions (e.g., large backgrounds) overshadow small objects, degrading training and rare-class performance.

Algorithmic strategies reweight decisions or features to improve rare-class detection. Chan et al.~\cite{chan2019application} use localized maximum-likelihood rules to reweight pixel predictions. Remote sensing’s multi-scale, complex scenes further motivate scale-adaptive designs: Wang et al.~\cite{wang2024unbalanced} fuse multi-scale features for unbalanced classes, and Zhou et al.~\cite{zhou2023dynamic} use effective-sample-based dynamic class balancing.

Long-tailed distributions are acute in outdoor LiDAR, where sparse, uneven sampling underrepresents small/dynamic classes. Architectures that preserve fine structure help: SPVCNN~\cite{tang2020searching} adds a high-resolution point branch to sparse voxels, while Cylinder3D~\cite{zhu2020cylindrical} and DS-Net~\cite{hong2020lidar} use cylindrical voxelization to improve dynamic foregrounds on SemanticKITTI and nuScenes. Loss reweighting also helps and is often paired with temporal/structural priors; e.g., LiDAR-UDA mitigates structural mismatch via beam-aware subsampling and stabilizes pseudo-labels over time, improving rare human classes in Syn2Real~\cite{shaban2023lidar}.

Augmentation combats imbalance and improves generalization. Classical geometric/photometric transforms increase diversity~\cite{alomar2023data}; generative models~\cite{ho2020denoising,rombach2022high} synthesize realistic samples with success in medical imaging~\cite{gadermayr2019generative} and underwater recognition~\cite{lin2023oysternet,lin2024odyssee}. Feature-space augmentation~\cite{li2021simple} generates novel samples from embeddings and is especially useful in semi-supervised regimes~\cite{kuo2020featmatch}.

LiDAR-specific mixing respects polar sampling and object geometry. PolarMix swaps azimuth sectors and applies instance-level rotations to enrich minorities, consistently boosting segmentation/detection~\cite{xiao2022polarmix}. CoSMix composes labeled synthetic with unlabeled real clouds for UDA~\cite{saltori2022cosmix}. For Syn2Real, SynLiDAR provides synthetic sequences and a Point-Cloud Translator to disentangle appearance/sparsity gaps, yielding stronger UDA/SSDA baselines~\cite{xiao2022transfer}. We adopt PolarMix for LiDAR and pair it with feature-space augmentation to densify rare-class manifolds.

Hyperbolic neural networks often struggle with few-shot generalization. HypMix~\cite{sawhney2021hypmix} addresses this by extending mixup into hyperbolic space: inputs are mapped to hyperbolic space, interpolated, and then mapped back to Euclidean space. To handle unlabeled data, the authors propose Möbius Gyromidpoint Label Estimation, which generates augmented inputs, maps their logits into hyperbolic space, interpolates them, and converts the result back to Euclidean space to form pseudolabels. These are then combined with labeled data for HypMix training.

Hyperbolic Feature Augmentation (HFA) tackles data scarcity by synthesizing class-identity-preserving features via per-class wrapped normal distributions on a hyperbolic manifold. To estimate each distribution’s curvature, mean, and covariance, HFA adopts a neural-ODE meta-learning scheme that treats parameter updates as a continuous gradient flow solved with RK4~\cite{butcher1996history}, leveraging prior structure to obtain accurate estimates in low-data regimes. A Euclidean upper bound on the augmentation loss further removes the need for expensive hyperbolic operations during training, while still enabling a distance-based classifier in hyperbolic space. Building on HFA, our RAFT framework extends these ideas from classification to segmentation and integrates complementary augmentations and active selection to explicitly address class imbalance and uncertainty, yielding stronger generalization and robustness in real-world deployments.

\section{Method}

In this section, we introduce RAFT (Robust Augmentation of FeaTures), our generalized framework for domain adaptation across both RGB image segmentation and LiDAR point cloud segmentation. We first provide an overview of Hyperbolic Active Learning Optimization (HALO) and Annotator's Voxel Confusion Degree (VCD), which form the foundation of our approach. We then present our extensions: a pixel/point-level adaptation of Hyperbolic Feature Augmentation (HFA), and the domain mixing strategies we adopt: Domain Adaptation via Cross-Domain Mixed Sampling (DACS) for RGB and PolarMix for LiDAR. Taken together, these components form a solution to the Syn2Real problem across RGB and LiDAR semantic segmentation.

\subsection{Hyperbolic Active Domain Adaptation}

Active domain adaptation (ADA) only acquires a handful of target labels per round, thus it aims to spend that budget where it removes the most uncertainty. For RGB, we build upon HALO, with its per-pixel acquisition score that multiplies the hyperbolic radius of each embedding - used as a data-scarcity indicator - with the prediction entropy, yielding a proxy for epistemic uncertainty that has proven effective for ADA segmentation under shift.

For LiDAR, we follow Annotator’s online, voxel-centric strategy and retain the Voxel Confusion Degree (VCD) to capture local class diversity/disagreement. We enrich this with a HALO-esque hyperbolic component by averaging radius$\times$entropy over points (or sparse-tensor features) within each candidate voxel, yielding a more informative LiDAR acquisition score without changing the querying scheme.


\subsection{Hyperbolic Feature Augmentation}

A key challenge in adapting HFA from image classification to dense prediction tasks lies in the complexity of the feature spaces. Generating detailed feature representations while preserving accurate spatial information is beyond the capabilities of the original neural ODE approach used in HFA.

Therefore, we take a different approach. While we still generate an approximate hyperbolic wrapped normal distribution for each semantic class via neural ODEs, we sample individual embeddings (pixel embeddings for RGB, voxel embeddings for LiDAR) from these class-specific distributions. We then perform weighted interpolation in hyperbolic space between the embeddings extracted from the data and our generated embeddings on a per-class basis.

For RGB segmentation, we sample 5 synthetic embeddings per class and interpolate with the extracted pixel embeddings. Due to the increased computational requirements of training LiDAR segmentation models, we sample only 2 synthetic embeddings per class to limit overhead.

We utilize the weighted M\"obius gyromidpoint~\cite{Ungar2009-pw}:
\begin{equation}
m_{\kappa}(x_{1}...x_{n},\alpha_{1}...\alpha_{n})= \frac{1}{2}\otimes_{\kappa}\left(\sum_{i=1}^{n}\frac{\alpha_{i}\lambda_{x_{i}}^{\kappa}}{\sum_{j=1}^{n}\alpha_{j}(\lambda_{x_{j}}^{\kappa}-1)}x_{i}\right)
\end{equation}
where \(x_{i}\) represents the real and sampled embeddings, \(\alpha_{i}\) is the weight, and \(\kappa\) is the curvature of the hyperbolic space, which we fix at -1.

To balance diversity and stability, we dynamically adjust the interpolation weights during training, gradually increasing the influence of the sampled embeddings. We forego the prototype classifier utilized in HFA and instead use a hyperbolic multinomial logistic regression classifier~\cite{ganea2018hyperbolic} for both modalities.

\subsubsection{Hyperbolic Mixup}

To further increase feature diversity in RGB segmentation, we optionally implement mixup in hyperbolic space. For each class, we take real pixel embeddings $\{h_i\}$, create shuffled pairs $\{h'_i\}$, sample mixing coefficients $\lambda_{i}\sim \text{Beta}(\alpha,\alpha)$, and perform geodesic interpolation using the M\"obius gyromidpoint:
\begin{equation}
\tilde{h}_{i}^{\text{mix}}=m_{\kappa}(h_{i},h_{i}^{\prime},\lambda_{i},1-\lambda_{i})
\end{equation}
We combine these mixed embeddings with the sampled embeddings from the neural ODE distributions into a single augmentation pool. When reintegrating features into the spatial feature map, we randomly select from either mixed or sampled embeddings on a per-class basis.

\subsubsection{Meta-Learning and ODE Integration}

We use a meta-learning approach to train the gradient flow networks for distribution estimation. For both RGB and LiDAR segmentation, we partition the source dataset into training and validation sets, use the training set to estimate distribution parameters via neural ODEs and train the segmentation model with generated augmentations, then evaluate on the validation set to update the gradient flow networks.

A key difference in our implementation lies in the neural ODE solver. For RGB segmentation, we use the adaptive RK4~\cite{butcher1996history} method to solve the continuous gradient flow. For LiDAR segmentation, due to the significantly larger point clouds (often 100k+ points per scan) and the computational demands of sparse 3D convolution networks, we use a simpler fixed-step Euler integration method with $\Delta t = 0.5$ over 2 steps, as it reduces memory and computational demands. For both modalities, we additionally use focal losses rather than standard classification losses, when evaluating whether the neural network receiving feature space augmentations is correctly classifying points.

The complete HFA loss is formulated as:
\begin{equation}
\begin{aligned}
\mathcal{L}_{\text{hfa}} =\ & 
\underbrace{\mathcal{L}_{\text{orig\_cls}} + \mathcal{L}_{\text{aug\_cls}}}_{\text{Classification Losses}} + \underbrace{\lambda_{\text{div}} \mathcal{L}_{\text{div}}}_{\text{Diversity Loss}} \\
& + \underbrace{\lambda_{\text{proto\_reg}} \mathcal{L}_{\text{proto\_reg}}}_{\text{Prototype Regularization}}  + \underbrace{\lambda_{\text{mean\_var}} \mathcal{L}_{\text{mean\_var}}}_{\text{Distribution Regularization}}
\end{aligned}
\end{equation}

\subsection{Domain Mixing Strategies}

The final component of our RAFT framework involves modality-specific domain mixing strategies to strengthen the feature extractor's representations.

\subsubsection{DACS for RGB Images}

For RGB images, we employ Domain Adaptation via Cross-Domain Mixed Sampling (DACS). We slightly modify it to leverage the certainty measures from HALO to identify high-confidence regions in the target domain, generate pseudo-labels for them, and then use a cut-and-paste approach to mix these regions with source domain images.

\subsubsection{PolarMix for LiDAR Point Clouds}

For LiDAR data, the sparse and irregular nature of point clouds means that geometry and spatial proximity plays a significantly greater role in semantic understanding. Therefore, rather than adapting DACS to LiDAR, we instead adopted PolarMix~\cite{xiao2022polarmix} without any additional modification. PolarMix applies an instance-wise cut, rotate, and paste (with pseudolabels) from the target to the source domain.

\subsection{Training Process}

In both modalities, we pretrain the segmentation model on a source dataset, then apply the RAFT framework. The primary difference between the modalities is in the application of domain mixing and HFA. The RGB training pipeline applies both domain mixing and HFA on every iteration, making the combined loss be:

\begin{equation}
\mathcal{L}_{\text{total}}^{\text{RGB}} = \mathcal{L}_{\text{src}} + \mathcal{L}_{\text{tgt}} + \lambda_{\text{hfa}}\mathcal{L}_{\text{hfa}} + \mathcal{L}_{\text{dacs}}
\end{equation}

For LiDAR segmentation, we alternate between applying HFA or domain mixing on each training iteration (up until the model begins training on the target domain exclusively). Thus the composite loss for LiDAR is:
\begin{equation}
\mathcal{L}_{\text{total}}^{\text{LiDAR}} = \mathcal{L}_{\text{src}} + \mathcal{L}_{\text{tgt}} + (\lambda_{\text{hfa}}\mathcal{L}_{\text{hfa}} \textbf{ or } \mathcal{L}_{\text{polarmix}})
\end{equation}
where $\lambda_{\text{hfa}} = 0.1$ for both modalities.
\section{Experimental Setup}

In this section, we detail our experimental protocol, including the datasets, model architectures, implementation details, and the robotics platform used for real-world validation.

\subsection{Datasets and Benchmarks}

We evaluate RAFT in both image and LiDAR semantic segmentation to measure Syn2Real transfer. For images, we adapt from the synthetic SYNTHIA~\cite{ros2016synthia} and GTA-V~\cite{Richter_2016_ECCV} sources to the real-world Cityscapes~\cite{cordts2016cityscapes} target. For LiDAR, we transfer from SynLiDAR~\cite{xiao2022transfer} to the real-world SemanticKITTI~\cite{behley2019iccv, behley2021ijrr, geiger2012cvpr} and SemanticPOSS~\cite{pan2020semanticposs} targets. These benchmarks are standard in autonomous driving and provide the synthetic $\rightarrow$ real shifts our method is designed to address.

\subsection{Implementation Details}

All our experiments are implemented in PyTorch. We leverage the geoopt~\cite{geoopt2020kochurov} library for hyperbolic geometry computations and torchdiffeq~\cite{torchdiffeq} for training the neural ODEs used in our Hyperbolic Feature Augmentation (HFA) module in image segmentation.

Within the image segmentation task the SegFormer (B4 variant)~\cite{xie2021segformer} and DeepLabv3+~\cite{chen2018encoder} architectures. For all image-based experiments, we resize images from GTA-V and SYNTHIA to $1280 \times 720$ and from Cityscapes to $1280 \times 640$. We use the AdamW optimizer with a base learning rate of $6 \times 10^{-5}$ and a polynomial learning rate schedule. For the HFA components, we use a smaller learning rate of $6 \times 10^{-6}$ without a scheduler. 

In LiDAR domain adaptation, as with  Annotator~\cite{xie2023annotator} we employ the MinkNet~\cite{choy20194d} and SPVCNN~\cite{tang2020searching} architectures. Due to the sparse nature of LiDAR point clouds, we make use of the TorchSparse~\cite{MLSYS2022_c48e8203} library to implement these models. For training we use the SGD optimizer with a base learning rate of 0.01, a weight decay of 0.0001, and a momentum of 0.9. We also use a linear warmup with a cosine decay learning rate scheduler. Additionally, unless specified otherwise, we used the default 5\% annotation budget of HALO, and Annotator's top-1 selected voxel per scan, repeated 5 separate times during training.

\subsection{Evaluation Metrics}

We evaluate the performance of our segmentation models using the standard metric of mean Intersection-over-Union (mIoU). For the SYNTHIA to Cityscapes benchmark, we report both the mIoU over 13 common classes (mIoU) and over all 16 classes (mIoU*), following the standard evaluation protocol. For all other benchmarks, we report a single mIoU score over all classes. As with Annotator, we utilize a reduced class mapping to align the classes from SynLiDAR to SemanticKITTI and SemanticPOSS datasets.

\subsection{Robotics Platform and Real-World Validation}

To demonstrate real-world applicability, we deploy our domain-adapted segmentation models on a Unitree Go2 Edu using a top-mounted Livox Mid-360 solid-state LiDAR and the Go2’s front-facing RGB camera. We recorded ROS bags while walking the robot through suburban outdoor areas; because this setup introduced significant domain shift (especially for LiDAR, and to a lesser extent RGB) relative to the datasets we benchmarked with, we annotated a small subset of collected point clouds~\cite{behley2019iccv} and RGB images to further adapt the models to the Go2’s sensors.


\section{Results}
In this section, we present the outcomes of our experiments across both RGB image and LiDAR point cloud semantic segmentation benchmarks, demonstrating RAFT's effectiveness as a unified domain adaptation framework.

\subsection{Quantitative Comparison}

\begin{table*}[htbp]
\centering
\footnotesize
\resizebox{\textwidth}{!}{
\begin{tabular}{llllllllllllllllllll}
\hline
Model                 & mIoU & mIoU* & \rotatebox{90}{road} & \rotatebox{90}{sidewalk} & \rotatebox{90}{building} & \rotatebox{90}{wall} & \rotatebox{90}{fence} & \rotatebox{90}{pole} & \rotatebox{90}{light} & \rotatebox{90}{sign} & \rotatebox{90}{vegetation} & \rotatebox{90}{sky}  & \rotatebox{90}{person} & \rotatebox{90}{rider} & \rotatebox{90}{car}  &   \rotatebox{90}{bus}  &   \rotatebox{90}{motorcycle} & \rotatebox{90}{bicycle} \\ 
\hline
\textbf{RAFT SegFormer B4 (ours)} & \textbf{79.9} & \textbf{83.5} & 98.3 & 87.1 & 93.0 & 66.1 & 64.6 & 62.2 & 69.2 & 77.8 & 93.3 & 95.2 & 81.8 & 62.9 & 95.4 & 89.2 & 65.9 & 76.8        \\
HALO SegFormer B4~\cite{franco2023hyperbolic}     & 77.8 & 82.1 & 98.3 & 86.5     & 92.6     & 61.0 & 61.5  & 60.6 & 67.6  & 76.2 & 93.2       & 94.6 & 80.8   & 58.9  & 95.0 &  85.1 & 62.7       & 75.6    \\
\textbf{RAFT DeepLabv3+ (ours)} & \textbf{76.9} & \textbf{81.7} & 98.0 & 84.7 & 92.0 & 55.0 & 54.3 & 59.7 & 66.1 & 75.9 & 92.7 & 94.6 & 79.9 & 59.4 & 94.9 & 86.4 & 62.1 & 75.4    \\
HALO DeepLabv3+~\cite{franco2023hyperbolic} & 75.6 & 80.2 & 97.5 & 81.5 & 91.5 & 56.5 & 52.7 & 57.0 & 63.2 & 72.9 & 92.0 & 94.4 & 77.8 & 57.4 & 94.4 & 86.1 & 60.5 & 73.5   \\
RIPU DeepLabv2~\cite{xie2022towards} & 70.1 & 75.7 & 96.8 & 76.6 & 89.6 & 45.0 & 47.7 & 45.0 & 53.0 & 62.5 & 90.6 & 92.7 & 73.0 &  52.9 & 93.1 & 80.5 &  52.4 & 70.1        \\
ILM-ASSL DeepLabv3+~\cite{guan2023iterative} & 76.6 & 82.1 & 97.4 & 80.1 & 91.8 & 38.6 & 55.2 & 64.1 & 70.9 & 78.7 & 91.6 & 94.5 & 82.7 & 60.1 & 94.4 & 81.7 & 66.8 & 77.2 \\
DWBA-ADA DeepLabv3+~\cite{dwba_ada2024} & 72.7 & 78.1 & 97.4 & 90.3 & 47.2 & 47.9 & 53.4 & 57.2 & 67.6 & 91.7 & 94.2 & 76.2 & 55.0 & 93.8 & 83.4 & 55.1 & 72.1 & 78.1 \\\hline
\end{tabular}
}
\vspace{1mm}
\captionsetup{font={footnotesize},labelfont=bf}
\caption{Comparison of Syn2Real methods for image segmentation on SYNTHIA to Cityscapes. mIoU* utilizes 13 classes, excluding "wall", "fence", and "pole", while mIoU utilizes all 16 classes within SYNTHIA.}
\vspace{-4mm}
\label{tab:synthia_results}
\end{table*}

\begin{table*}[htbp]
\centering
\footnotesize
\resizebox{\textwidth}{!}{
\begin{tabular}{lllllllllllllllllllll}
\hline
Model                 & mIoU & \rotatebox{90}{road} & \rotatebox{90}{sidewalk} & \rotatebox{90}{building} & \rotatebox{90}{wall} & \rotatebox{90}{fence} & \rotatebox{90}{pole} & \rotatebox{90}{light} & \rotatebox{90}{sign} & \rotatebox{90}{vegetation} & \rotatebox{90}{terrain} & \rotatebox{90}{sky}  & \rotatebox{90}{person} & \rotatebox{90}{rider} & \rotatebox{90}{car}  &  \rotatebox{90}{truck}  &   \rotatebox{90}{bus}  &  \rotatebox{90}{train}  &   \rotatebox{90}{motorcycle} & \rotatebox{90}{bicycle} \\ \hline
\textbf{RAFT SegFormer B4 (ours)} & \textbf{78.2} & 98.3 & 85.8 & 92.7 & 63.8 & 62.7 & 61.6 & 69.2 & 77.3 & 92.5 & 64.0 & 94.9 & 80.9 & 62.3 & 95.1 & 86.5 & 86.1 & 73.4 & 63.3 & 75.6          \\
HALO SegFormer B4~\cite{franco2023hyperbolic}     & 77.8 & 98.2 & 85.4     & 92.5     & 62.5 & 61.6  & 58.3 & 67.7  & 74.9 & 92.2       & 65.1 & 94.7   & 79.9  & 60.8 &  94.6 & 84.1       & 85.4 & 83.6 & 61.2 & 75.5    \\
\textbf{RAFT DeepLabv3+ (ours)} & \textbf{74.8} & 97.9 & 83.4 & 92.0 & 56.2 & 56.1 & 59.1 & 65.1 & 74.7 & 91.9 & 63.7 & 94.5 & 79.2 & 58.9 & 94.3 & 77.7 & 81.2 & 58.0 & 62.4 & 74.2          \\
HALO DeepLabv3+~\cite{franco2023hyperbolic}     & 74.5 & 97.6 & 81.0     & 91.4     & 53.7 & 54.9  & 56.7 & 62.9  & 72.1 & 91.4       & 60.5 & 94.1   & 78.0  & 57.3 &  94.0 & 81.4       & 84.7 & 70.1 & 60.0 & 73.3    \\
RIPU DeepLabv2~\cite{xie2022towards} & 71.2 & 97.0 & 77.3 & 90.4 & 54.6 & 53.2 & 47.7 & 55.9 & 64.1 & 90.2 & 59.2 & 93.2 &  75.0 & 54.8 & 92.7 &  73.0 & 79.7 & 68.9 & 55.5 & 70.3        \\
ILM-ASSL DeepLabv3+~\cite{guan2023iterative} & 76.1 & 96.9 & 77.8 & 91.6 & 46.7 & 56.0 & 63.2 & 70.8 & 77.4 & 91.9 & 54.9 & 94.5 & 82.3 & 61.2 & 94.9 & 79.3 & 88.1 & 75.3 & 65.8 & 77.6 \\
DWBA-ADA DeepLabv3+~\cite{dwba_ada2024} & 71.9 & 97.5 & 80.5 & 90.8 & 54.7 & 52.2 & 53.3 & 55.7 & 65.2 & 91.0 & 61.0 & 93.5 & 75.3 & 53.6 & 92.9 & 81.8 & 75.2 & 62.9 & 57.8 & 71.6 \\ \hline
\end{tabular}
}
\vspace{1mm}
\captionsetup{font={footnotesize},labelfont=bf}
\caption{Comparison of Syn2Real methods for image segmentation on GTAV to Cityscapes}
\vspace{-4mm}
\label{tab:gtav_results}
\end{table*}
\vspace{-2mm}

\begin{table*}[htbp]
\centering
\footnotesize
\resizebox{\textwidth}{!}{
\begin{tabular}{lllllllllllllllllllll}
\hline
Model                        & mIoU          & \rotatebox{90}{car}  & \rotatebox{90}{bicycle} & \rotatebox{90}{motorcycle} & \rotatebox{90}{truck} & \rotatebox{90}{other vehicle} & \rotatebox{90}{person} & \rotatebox{90}{bicyclist} & \rotatebox{90}{motorcyclist} & \rotatebox{90}{road} & \rotatebox{90}{parking} & \rotatebox{90}{sidewalk} & \rotatebox{90}{other-ground} & \rotatebox{90}{building} & \rotatebox{90}{fence} & \rotatebox{90}{vegetation} & \rotatebox{90}{trunk} & \rotatebox{90}{terrain} & \rotatebox{90}{pole} & \rotatebox{90}{sign} \\ \hline
\textbf{RAFT MinkNet (ours)} & \textbf{59.6} & 95.0 & 29.1    & 60.0       & 72.9  & 42.5          & 69.5   & 80.2      & 6.4          & 80.6 & 30.7    & 77.0     & 2.6          & 91.3     & 66.2  & 74.0       & 68.4  & 77.0    & 63.2 & 45.0         \\
Annotator MinkNet            & 56.3          & 94.3 & 0.9     & 60.8       & 61.2  & 44.5          & 63.7   & 76.6      & 3.4          & 87.1 & 34.4    & 71.4     & 2.8          & 86.2     & 62.0  & 86.2       & 60.4  & 71.7    & 59.6 & 42.2         \\
\textbf{RAFT SPVCNN}         & \textbf{56.8} & 94.6 & 21.4    & 63.1       & 64.0  & 44.7          & 61.9   & 72.4      & 0.3          & 88.0 & 28.3    & 75.1     & 2.7          & 87.3     & 56.7  & 85.5       & 64.9  & 69.9    & 57.9 & 40.6         \\
Annotator SPVCNN             & 55.7          & 95.0 & 19.4    & 60.1       & 49.5  & 44.5          & 59.6   & 75.4      & 1.1          & 87.8 & 30.4    & 74.6     & 2.3          & 84.8     & 53.6  & 85.4       & 62.8  & 71.7    & 60.8 & 38.7           \\ \hline
\end{tabular}
}
\vspace{1mm}
\captionsetup{font={footnotesize},labelfont=bf}
\caption{Comparison of Syn2Real methods for LiDAR segmentation on SynLiDAR to SemanticKITTI}
\vspace{-4mm}
\label{tab:syn2kitti_results}
\end{table*}

\begin{table*}[htbp]
\centering
\footnotesize
\resizebox{\textwidth}{!}{
\begin{tabular}{lllllllllllllll}
\hline
Model                        & mIoU          & \rotatebox{90}{car}  & \rotatebox{90}{bicycle} & \rotatebox{90}{person} & \rotatebox{90}{rider} & \rotatebox{90}{ground} & \rotatebox{90}{building} & \rotatebox{90}{fence} & \rotatebox{90}{plants} & \rotatebox{90}{trunk} & \rotatebox{90}{pole} & \rotatebox{90}{sign} & \rotatebox{90}{bin} & \rotatebox{90}{cone} \\ \hline
\textbf{RAFT MinkNet (ours)} & \textbf{49.9} & 74.3 & 31.3    & 66.7   & 57.9  & 73.4   & 61.8     & 66.6  & 54.0   & 48.7  & 36.2 & 16.9         & 22.9        & 38.1       \\
Annotator MinkNet            & 47.8          & 53.5 & 32.0    & 53.0   & 57.0  & 59.6   & 66.7     & 54.3  & 63.5   & 40.0  & 34.2 & 37.3         & 31.6        & 38.5       \\
\textbf{RAFT SPVCNN (ours)}  & 48.2          & 65.0 & 32.5    & 66.1   & 56.7  & 82.2   & 76.0     & 51.8  & 76.3   & 41.9  & 29.4 & 16.3         & 23.1        & 15.1       \\
Annotator SPVCNN             & \textbf{49.7} & 68.2 & 32.2    & 68.3   & 57.6  & 76.9   & 69.9     & 57.1  & 70.8   & 38.8  & 33.8 & 14.8         & 19.9        & 38.5       \\ \hline
\end{tabular}
}
\vspace{1mm}
\captionsetup{font={footnotesize},labelfont=bf}
\caption{Comparison of Syn2Real methods for LiDAR segmentation on SynLiDAR to SemanticPOSS}
\vspace{-4mm}
\label{tab:syn2poss_results}
\end{table*}

Across both modalities, RAFT demonstrates improved performance over our baseline active learning-only methods. Table~\ref{tab:synthia_results} shows the results of SYNTHIA $\rightarrow$ Cityscapes domain adaptation. RAFT's performance exceeds that of other state-of-the-art active learning-only methods across both RGB image segmentation architectures. With SegFormer B4, RAFT achieves a 13-class mIoU of 79.9\% and a 16-class mIoU of 83.5\%, representing improvements of 2.1\% and 1.4\% respectively over HALO. With DeepLabv3+, RAFT achieves 76.9\% mIoU and 81.7\% mIoU*, or +1.3\% and +1.5\% respectively over HALO using the same architecture and annotation budget. On GTAV $\rightarrow$ Cityscapes (Table~\ref{tab:gtav_results}), RAFT still shows consistent improvements across both architectures. With SegFormer B4, RAFT achieves 78.2\% mIoU, a 0.4\% gain over HALO, while with DeepLabv3+, RAFT achieves 74.8\% mIoU, a 0.3\% improvement. 

In LiDAR segmentation, RAFT still performs quite well, although interestingly there is a regression on one benchmark instance. As shown in Table~\ref{tab:syn2kitti_results}, on the SynLiDAR $\rightarrow$ SemanticKITTI benchmark, RAFT achieves an mIoU of 59.6\% using MinkNet, a 3.3\% improvement over the Annotator baseline. With SPVCNN, RAFT achieves an mIoU of 56.8\%, a 1.1\% improvement. On the SynLiDAR $\rightarrow$ SemanticPOSS benchmark (Table~\ref{tab:syn2poss_results}), the MinkNet model trained using RAFT achieves a 49.9\% mIoU, a 2.1\% improvement over Annotator. However, the RAFT-trained SPVCNN model regresses from the Annotator-only trained SPVCNN model, exhibiting a 1.5\% degradation in mIoU.
As we will elaborate on in the following subsection, this degradation appears to largely be the fault of HFA being unable to properly model the distributions for certain classes in SemanticPOSS. Even in MinkNet - where RAFT demonstrates improvement over Annotator: much of the improvement in mIoU appears to stem from large IoU improvements on large and/or static geometry (building, ground, car), offsetting degradation on small, thin, and/or dynamic classes (plants, signs, bins). For SPVCNN, the classes overall largely exhibit on-par or slightly worse IoU's, with significantly lower IoU's on fences and cone/stones dropping the mIoU below that of Annotator.

\subsection{RAFT Component Ablation}

Naturally, given that RAFT applies several techniques over top of plain active learning, an obvious question is how much each component actually contributes to the overall result. Therefore we performed an ablation study for RGB and LiDAR segmentation with RAFT.

\begin{figure}
  \includegraphics[width=\linewidth]{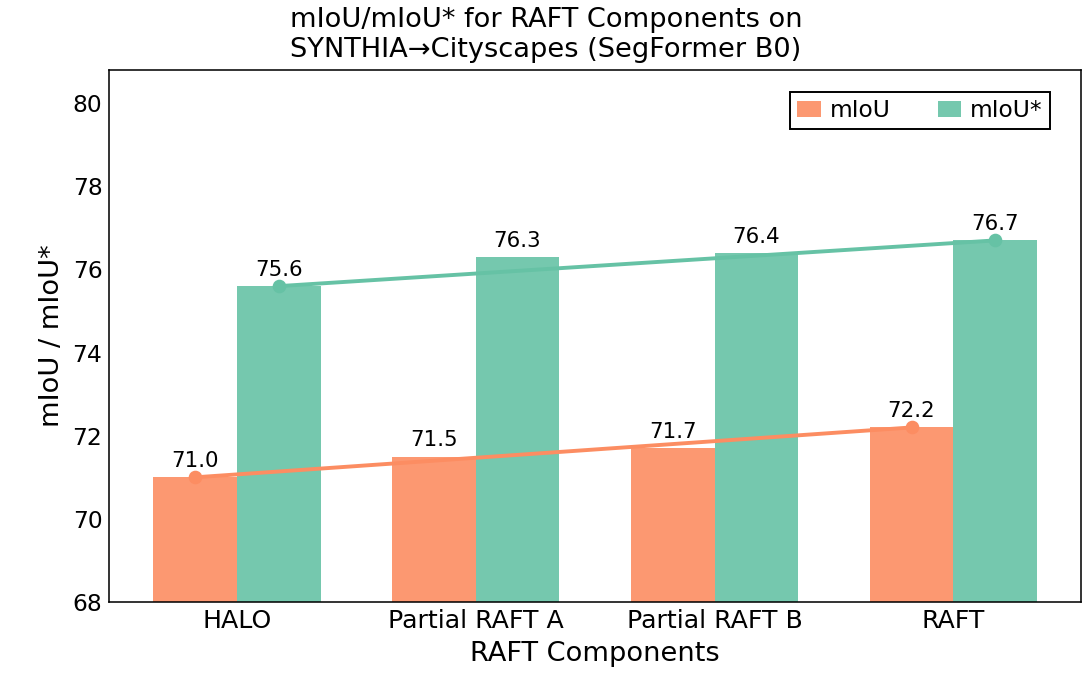}
  \captionsetup{font={footnotesize},labelfont=bf}
  \caption{The effect on mIoU and mIoU* of applying various RAFT components in performing domain adaptation of a SegFormer B0 model from SYNTHIA to Cityscapes. The mIoU* metric uses 13 common classes in both SYNTHIA and Cityscapes, while the mIoU metric uses all 16 classes shared between SYNTHIA and Cityscapes. Partial RAFT A includes HALO along with HFA and hyperbolic mixup, Partial RAFT B includes the aforementioned components plus the focal loss, and RAFT includes all RAFT components.}
  \label{fig:raft_ablation}
\end{figure}

As shown in Figure \ref{fig:raft_ablation}, each component plays a role in the final RAFT mIoU upon domain transfer. Using the SegFormer B0 architecture trained via HALO as our baseline, we achieve an mIoU and mIoU* of 71.0\% and 75.6\% respectively on SYNTHIA $\rightarrow$ Cityscapes. 

Adding HFA and hyperbolic mixup (Partial RAFT A) results in an mIoU and mIoU* of 71.5\% and 76.3\% respectively, representing a 0.5\% and 0.9\% improvement over HALO alone. Further extending with focal loss (Partial RAFT B) achieves 71.7\% and 76.4\% mIoU and mIoU*, a modest 0.2\% and 0.1\% improvement over Partial RAFT A. Finally, integrating DACS to form the complete RAFT framework results in an mIoU and mIoU* of 72.2\% and 76.7\% respectively, a 0.5\% and 0.3\% improvement over Partial RAFT B.

\begin{table}
\centering
\footnotesize
\resizebox{1.0\linewidth}{!}{
\begin{tabular}{llllll}
\hline
\multicolumn{2}{l}{}                       & \multicolumn{2}{l}{Syn to KITTI} & \multicolumn{2}{l}{Syn to POSS} \\ \cline{3-6} 
ADA Method                & Components     & MinkNet         & SPVCNN         & MinkNet        & SPVCNN         \\ \hline
HALO                      & n/a            & 54.3            & 55.6           & 39.9           & 47.7           \\ \hline
\multirow{2}{*}{VCD}      & n/a            & 56.3            & 55.6           & 47.8           & 49.7           \\
                          & HFA + PolarMix & 59.1            & 54.1           & 46.2           & 49.4           \\ \hline
\multirow{4}{*}{HALO-VCD} & n/a            & 56.0            & 56.5           & 42.6           & 48.5           \\
                          & PolarMix       & 58.1            & 48.9           & 48.3           & \textbf{50.9}  \\
                          & HFA            & 58.0            & \textbf{56.8}  & 47.4           & 48.8           \\
                          & HFA + PolarMix & \textbf{59.6}   & \textbf{56.8}  & \textbf{49.9}  & 48.6           \\ \hline
\end{tabular}
}
\vspace{1mm}
\captionsetup{font={footnotesize},labelfont=bf}
\caption{The effect on mIoU of applying the various RAFT modifications and components adapating the LiDAR segmentation models across the two benchmarks.}
\vspace{-7mm}
\label{tab:raft_lidar_ablation}
\end{table}

For LiDAR, as we made changes to the active learning strategy in addition to the augmentations, we performed several more ablations than for RGB segmentation. Table \ref{tab:raft_lidar_ablation} displays the results for both SynLiDAR $\rightarrow$ SemanticKITTI and SynLiDAR $\rightarrow$  SemanticPOSS using MinkNet and SPVCNN. We use HALO to describe the active selection using prediction entropy and hyperbolic radius, as in the original HALO for image segmentation. HALO-VCD refers to the combination of the just-described HALO with VCD. 

As in the RGB study, most settings improve over their baselines. The notable exception is SPVCNN on SynLiDAR$\rightarrow$SemanticPOSS, where the best configuration is HALO–VCD with PolarMix only; adding HFA reduces mIoU. We hypothesize this stems from the SemanticPOSS sensor (Hesai Pandar40P~\cite{pandar40plidar}) and SPVCNN’s point branch. The Pandar40P has 40 vertical channels with non-uniform vertical spacing, yielding a higher point density near the central elevation and sparser sampling toward the extremes. This anisotropic sampling produces thin, fragmentary observations for small or slender objects. HFA attempts to expand class distributions in feature space, but under such sparse, uneven geometry it can synthesize embeddings that do not correspond to stable input-space neighborhoods, making the point branch prone to overfit sensor-specific artifacts. Consistent with this explanation, the largest drops when adding HFA occur on small/thin classes (e.g., fence and cone/stone), where the uneven vertical sampling is most pronounced.

\subsection{Real-World Robot Validation}

\begin{figure}
  \includegraphics[width=1.0\linewidth]{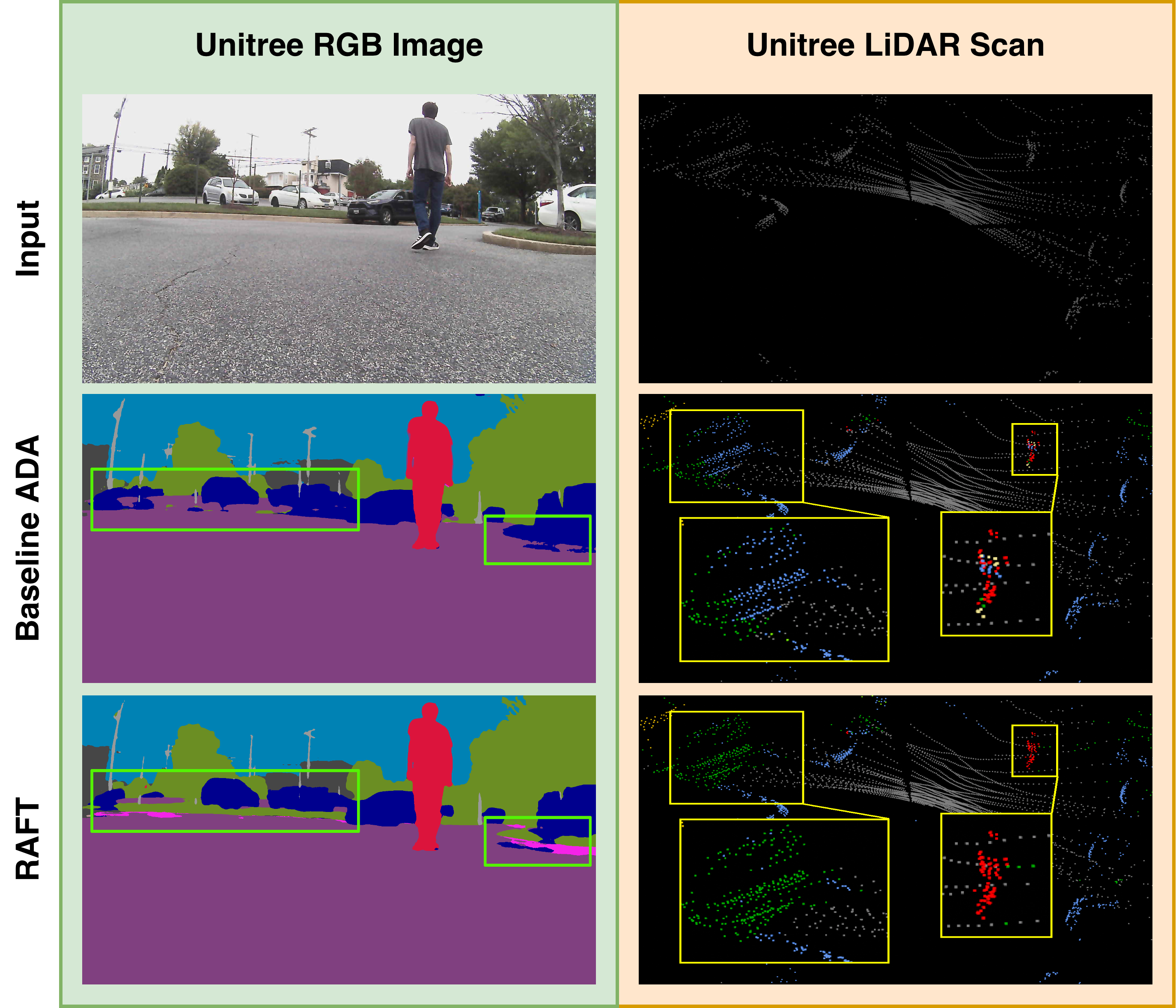}
  \captionsetup{font={footnotesize},labelfont=bf}
  \caption{The \textbf{Input} row contains the raw RGB and LiDAR scans collected from our Unitree Go2 Edu robot. The \textbf{Baseline ADA} row displays the segmentation masks created by the baseline ADA-only adapted models. Finally, the \textbf{RAFT} row displays the segmentation masks created by our RAFT-adapted models. Due to the sparsity of the LiDAR point clouds, we magnify the two regions which display improvements with the RAFT-trained model.}
  \vspace{-5mm}
  \label{fig:segmentation_results}
\end{figure}

Across the two modalities, both the active learning-only baseline and RAFT-trained models generally perform quite well at segmenting the recorded Go2 inputs, as shown in Figure \ref{fig:segmentation_results} and our associated video. The largest difference we noticed between the HALO and RAFT-trained DeepLabv3+ models is in classifying grass. The HALO-trained model effectively classified none of it properly, instead mistaking it as being part of a car. The RAFT-trained DeepLabv3+ does significantly better in this regard - with the model correctly classifying much of the grass as vegetation, albeit still struggling with identifying the curbs surrounding the grass. 

Despite the RAFT-trained MinkNet model exhibiting a larger increase in performance on our validation data over the Annotator-trained MinkNet model, the differences between the two models are somewhat subtle. The largest difference we noticed was in classifying the plants immediately in front of the building we recorded our demonstration data in front of. The Annotator-trained model mostly classified this vegetation as being part of a car, likely due to the cars on either side of it as seen in Figure \ref{fig:segmentation_results}. On the other hand, the RAFT-trained model, while not completely perfect, is largely able to classify the plants correctly as vegetation. 

One other small difference we noticed is that the RAFT-trained MinkNet is better at classifying people at longer distances compared to its Annotator counterpart. While both models fail once the people are simply too far away, Figure \ref{fig:segmentation_results} demonstrates that the Annotator model begins to struggle before the RAFT model does. One can imagine how this in particular is quite important in ensuring safe human-robot interaction, as the further away a person can be identified, the more time a robot has to slow down or change course before the risk of collision.

\begin{figure}
  \includegraphics[width=\linewidth]{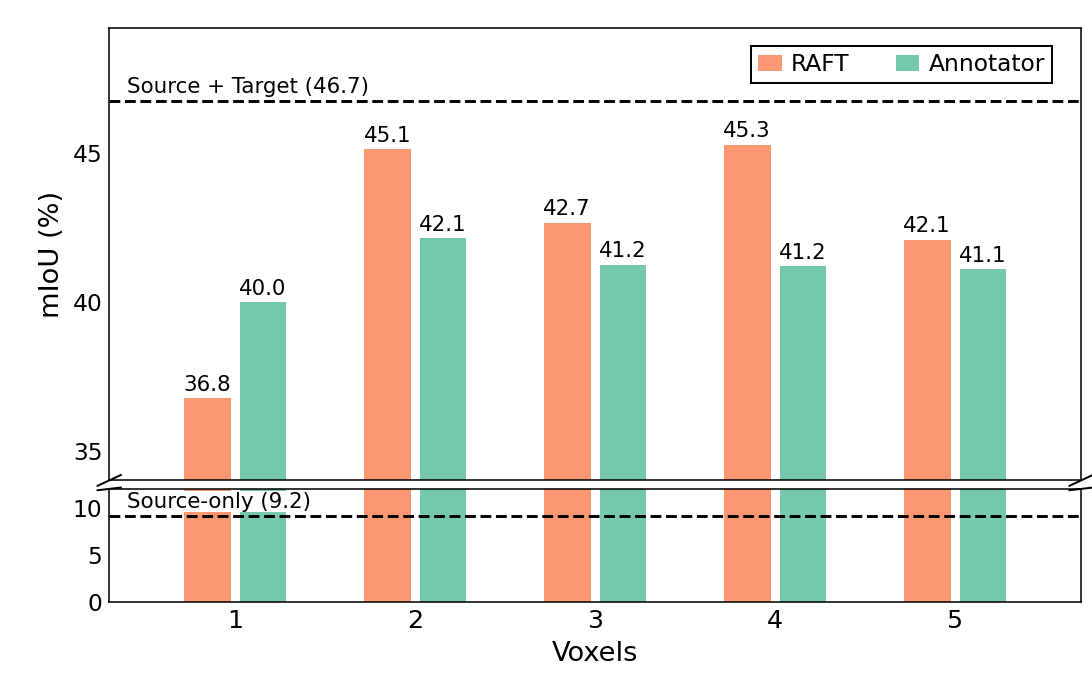}
  \captionsetup{font={footnotesize},labelfont=bf}
  \caption{The quantitative mIoU results of applying RAFT and Annotator with varying voxel budgets for active learning. While RAFT initially performs worse than Annotator; with the addition of only one more voxel for active learning, it not only outperforms Annotator, but comes close to matching the performance of the model trained on a mix of source SynLiDAR and target Livox Mid-360 LiDAR scans.}
  \vspace{-5mm}
  \label{fig:raft_unitree_voxel_miou}
\end{figure}


With a fixed 5\% annotation budget for domain adaptation on DeepLabv3+, RAFT yields a small but consistent gain over HALO on the validation set (mIoU/mIoU$^*$: 37.4\%/42.6\% for HALO vs.37.7\%/43.1\%  for RAFT), aligning with qualitative improvements on our collected data. For MinkNet, however, RAFT{+}Annotator initially lagged behind Annotator when using the same sparse LiDAR voxel budget (1 voxel/scan). Increasing the budget to 2 voxels/scan reversed this trend. RAFT surpassed Annotator, with the margin growing at higher budgets. We hypothesize that, given our small dataset, the ultra-sparse regime encouraged overfitting and spurious selections under RAFT; a modestly larger voxel budget provides a healthier mix of informative and noisy labels, mitigating overfitting and improving performance.
\subsection{Annotated Livox Mid-360 LiDAR Dataset}

We collected and fully labeled a sequence collected from the Go2 using SemanticPOSS' label scheme. The resulting dataset contains 200 scans for training, and 51 for testing. The data consists of a parking lot surrounded by trees, plants, some posts, one large foreground building and a few background buildings, as well as people and cars. Thus, the data does not use every single class available in SemanticPOSS. Despite the small size of the dataset, as demonstrated by our Go2 results, which were recorded in a separate area, this dataset is useful for finetuning and domain adaptation.





\section{Conclusion}

We presented RAFT, a framework for synthetic-to-real domain adaptation in RGB and LiDAR semantic segmentation. RAFT builds on active learning to make a fixed annotation budget go further by (i) exploiting hyperbolic geometry for selection and uncertainty calibration, (ii) extending hyperbolic feature augmentation from image classification to dense prediction via pixel/voxel embeddings, and (iii) applying modality-appropriate domain mixing (DACS for RGB, PolarMix for LiDAR). In effect, HFA expands the per-class distribution without labels, mixing improves representation alignment, and the active selector spends the budget on the residual, harder-to-anticipate cases (HALO for RGB; HALO–VCD for LiDAR).

Across SYNTHIA/GTAV $\rightarrow$ Cityscapes and SynLiDAR $\rightarrow$ SemanticKITTI/POSS, RAFT consistently improves over strong ADA baselines while using only a small fraction of target labels (e.g., +2.1 mIoU* with SegFormer-B4 on SYNTHIA $\rightarrow$ Cityscapes; +3.3 mIoU with MinkNet on SemanticKITTI), and it transfers to a Unitree Go2 with a Livox Mid-360, where we also release a small annotated sequence to support further research. Ablations show HFA delivers early gains and mixing provides the final push once features are partially aligned; for LiDAR, the best recipe depends on backbone and sensor sampling. Notably, SPVCNN on POSS benefits from PolarMix without HFA, suggesting a future research direction in improving HFA for LiDAR sensors with non-uniform vertical resolutions.

\bibliographystyle{IEEEtran}
\bibliography{references}


\end{document}